# Hindi to English Transfer Based Machine Translation System

**Akanksha Gehlot[1], Vaishali Sharma[2], Shashipal Singh[3], Ajai Kumar[4]**

[1, 2] Banasthali Vidyapith, Rajasthan India

[1]akankshagehlot2@gmail.com

[2]Vaishali.sharma2217@gmail.com

[3,4]AAIG, Centre for development of advanced computing, Pune India

[3]shashipalsingh@gmail.com

[4] ajai@cdac.in

## Abstract

*In large societies like India there is a huge demand to convert one human language into another. Lots of work has been done in this area. Many transfer based MTS have developed for English to other languages, as MANTRA CDAC Pune, MATRA CDAC Pune, SHAKTI IISc Bangalore and IIIT Hyderabad. Still there is a little work done for Hindi to other languages. Currently we are working on it. In this paper we focus on designing a system, that translate the document from Hindi to English by using transfer based approach. This system takes an input text check its structure through parsing. Reordering rules are used to generate the text in target language. It is better than Corpus Based MTS because Corpus Based MTS require large amount of word aligned data for translation that is not available for many languages while Transfer Based MTS requires only knowledge of both the languages(source language and target language) to make transfer rules. We get correct translation for simple assertive sentences and almost correct for complex and compound sentences.*

## 1. Introduction

A Machine Translation system essentially takes a text in one language (Source Language) and converts it into another language (Target Language). Source and target languages are natural languages as Hindi, English, as opposed to manmade languages as C or SQL. There are various approaches of machine translations Dictionary Based and Rule Based. In Dictionary Based Machine Translation, translation is based on entries of a language dictionary. The word's equivalent is used to develop the translation. Rule Based Machine Translation deals with morphological, syntactic and semantic information of source language and target language. It uses bilingual dictionaries and grammar rules. It has various approaches as Direct Approach, Transfer Based, Knowledge Based, Corpus Based, Context Based and Example Based. Here we are using Transfer Based approach in our project.

In Transfer Based System a source language is transformed into an abstract representation then its equivalent is generated for target language using bilingual dictionaries and grammar rules. This system has three major components: Analysis, Transfer and Synthesis.

Analysis of the source text is based on linguistic information such as morphology, Part of speech, syntax, semantic,etc. Algorithms are applied to parse the source language and derive

- The syntactic structure of the text to be Translated
- The semantic structure

Transfer module; transfer the source language structure into its equivalent target language structure. Some arrangement rules are used by which source language syntax structure can be transformed into target language syntax structure. Set of transfer rules are used for Hindi to English machine translation system are [5]

Hindi Structure English Structure

S ->NP VP            S->NP VP

NP->PRON NOUN    NP->PRON NOUN

VP->ADV Y           VP->AUX Y

Y->ADJ AUXY->ADV ADJ

Generation module; generate target language text using target language structure.

Syntactic and lexical ambiguities can be resolved better than direct translation approaches.

Example:

Input:यह किताब बहुत अच्छी है

Output:This book is very good

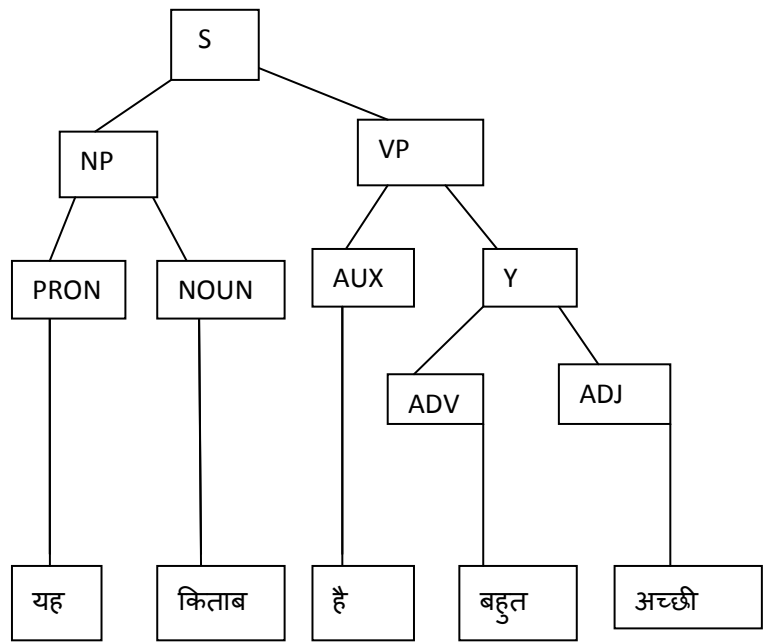

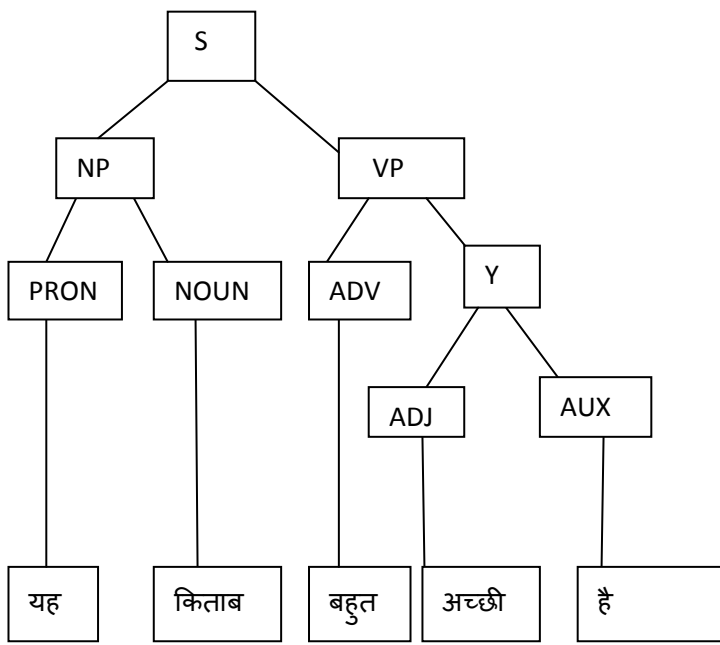

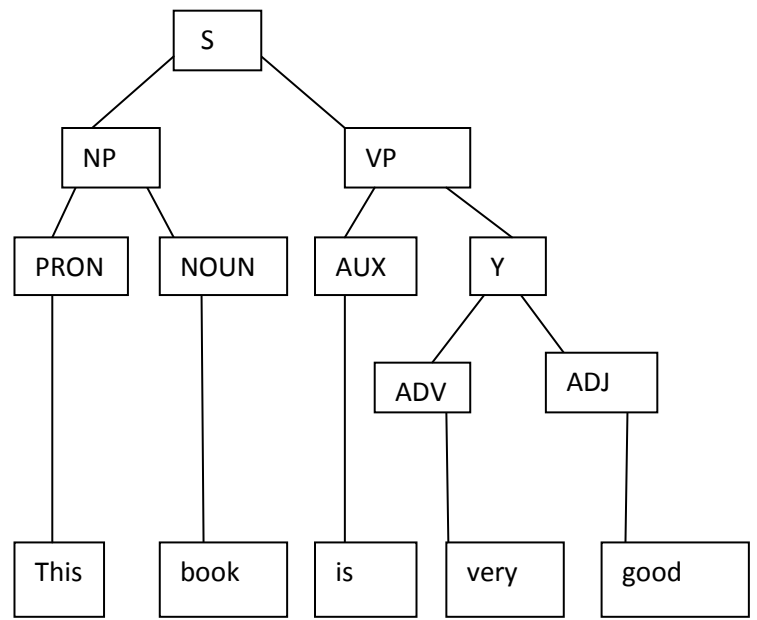

**Fig 1: Structural transfer in Hindi-English MTS**

We have used two transfer rules. First transform VP-> ADV Y into VP->AUX Y.Second, transform Y->ADJ AUX into Y->ADV ADJ.

Transfer system is more realistic and flexible to meet the needs of different level of syntactic and semantic analysis.

## 2. Related Work [5]

**Table 1: Existing transfer based MTS**

| Machine Translation System | Developer | Year |
|---|---|---|
| MANTRA-English to Hindi, Telugu, Gujarati | CDAC PUNE | 1995 |
| MATRA-English to Hindi MTS | CDAC PUNE | 2004 |
| SHAKTI-English to Hindi,Marathi,Telugu MTS (combines rule based and statistical approach) | IISc Bangalore and IIIT Hyderabad | 2004 |
| Anubaad-English-Bengali MTS(n-gram approach for pos tagging) | CDAC Kolkata | 2005 |
| English to Kannada MTS using UCSG | University of Hyderabad | 2006 |
| English – Malayalam MTS | Amrita Institute of Technology | 2009 |

### 2.1 Mantra MT System

MANTRA (Machine assisted translation tool) translates the document from English to Hindi in a specified domain of personal administration, office order, office memorandums and circulars. Mantra uses Tree Adjoining Grammar for parsing and generation. It takes input in the form of text or output of speech recognition program.[6]

### 2.2 MATRA MT System

MATRA, human-aided transfer based translation system for English to Hindi. The system has been applied mainly in the domains of news, technical phrases and annual reports. [6]

### 2.3 SHAKTI MT System

This system works on three languages as Hindi, Marathi, and Telugu. If the user is not satisfied with the translation then it ask for meaning of input text components as word, phrases and even sentence level from user and retranslate it. [6]

### 2.4 English to Kannada MT System

This system is based on Transfer Based approach and has been applied in the domain of government circulars. Now efforts are on applying it on English to Telugu.[6]

## 3. Structural Differences between Hindi and English sentences

In Hindi sentences object appears before verb. It follows SOV structure while in English sentences object appears after verb it follows SVO structure. [2]

Input: सीतालड़कीहै

Output : Sita is girl

In Hindi sentence preposition is considered as postposition and it appears after noun while in English structure it follows noun.

Input:जवाहरलालनेहरुभारतकेप्रथमप्रधानमंत्रीथे

Output: Jawahar Lal Nehru was first prime minister of India.

Adjectives appear before noun in both the languages as Hindi and English.

Input: वहबुद्धिमानलड़कीहै

Output: She is intelligent girl

Adverb appears before verb in Hindi sentences while in English sentences, it appears after verb.

Input:मोहनतेजदौड़ताहै

Output: Mohan runs fast

Auxiliary verb appears after main verb in Hindi structure while in English structure it appears before main verb.

Input:रियाकाफीपीरहीहै

Output: Riya is drinking coffee

## 4. Methodology

For parsing we have used CYK algorithm. It follows bottom up approach. It works on context free grammar. Context Free Grammar allows the construction of efficient parsing algorithm. Context Free Grammar is defined by 4 tuples.

G= (V, T, P, S),Where,

V->set of variables

T->set of terminals

P->production rule,

S->Start Symbol.

For CYK algorithm, context free grammar is converted into Chomsky Normal Form. This algorithm rules used in this algorithm are in the form A->BC,B->b, and C -> c.

### 4.1 Cocke Younger Kasami Algorithm

CYK(Cocke Younger Kasami) Algorithm is a parallel dynamic syntax analysis algorithm. A two dimension matrix :{ Z (p,q)}.

Among it: each element Z(p,q) according to non-terminal set of all possible forms of phrase with certain span of input sentences. [1]

Abscissa p: position of the first word on the left of the span.

Ordinate q: number of word span contained. For p=1 to n (filling in the first row length is p)

1)For every rule A->$R_i$, adding the non-terminals A to setZ (m,m);

2) For q=2 to n (filling in other rows, length is q )

For p=1….q -n+1 (for all start point p)

ForK=1……. q-1(for the length of all left child nodes is k)

**Basic ideas table filling algorithm**

Two dimensional matrix can be used to encode the structure of a tree. Each cell [p,q] in this matrix contain a set of non-terminal that represent all the content that span position p through q of the input. Cell that represent entire input reside in position [1, n] in the matrix. Since grammar used is in CNF, non terminal entries in the table have exactly two child in parse.[1]

| Z (1,5) | | | | |
|---|---|---|---|---|
| Z (1, 4) | Z (2, 5) | | | |
| Z (1, 3) | Z (2, 4) | Z (3, 5) | | |
| Z (1, 2) | Z (2, 3) | Z (3, 4) | Z (4, 5) | |
| Z (1, 1) | Z (2, 2) | Z (3, 3) | Z (4, 4) | Z (5,5) |
| R1 | R2 | R3 | R4 | R5 |

**Fig 2: Structure of table in table filling algorithm**

For each content represented by an entry [ p,q] in the table there must be a position in input k, where it can be split into two parts as p<k<q. First content

[p,k] must lie to the left of the entry[p,q] along row p second entry [k,q] must lie  beneath it, along column q.

| S | | | |
|---|---|---|---|
| VP | ¢ | | |
| NP | | VP | |
| PRON | NOUN | VP | AUX |
| वह | बाज़ार | जाती | है |

**Fig 3: Example of CYK Algorithm**

S -> NP VP

NP ->PRON NOUN

VP ->VP AUX

PRON->वह

NOUN->बाज़ार

VP->जाती

AUX->है

¢is placed where no rules are formed

**4.2SystemArchitecture**

**4.2.1 Preprocessing**

In preprocessing phase, numbers of operation are applied to input data to make it executable by translation system. It includes treatment of punctuation and special characters.

**4.2.2Tokenizer**

Tokenizer, which is also known as lexical analyzer or word segmenter segment a sentence into meaningful unit known as tokens. Tokenizer takes output of preprocessing phase as an input. Individual units or tokens are extracted and processed to find its corresponding meaning in target language. This module uses space and punctuation mark as delimiter, extract token one by one from input text and give it to translation system till the input text is processed.

Input: अब्दुल कलाम महान वैज्ञानिक है

Output:  Token1   अब्दुल

Token2   कलाम

Token3   महान

Token4   वैज्ञानिक

Token5   है

**4.2.3Hindi Morphological Analyzer**

Morphological analyzer identifies the root word and its corresponding category for tokens that are generated by Tokenizer. [4]

Input:सीता बहुत अच्छी लड़की है

Output:

सीता   NOUN

बहुत   ADV

अच्छी   ADJ

लड़की   NOUN

है   AUX

**4.2.4 Transliteration**

For proper noun, we use transliteration instead of translation. Words or tokens that are not found in our database are transliterated.

Input: अजय

Output: ajay

For transliteration we maintain a table in database, containing Hindi letters and matras. In above example

अ -a ज –ja य –ya

If character at the end of word is not equal to "ा" then delete 'a' from its English meaning. English meaning of 'य' is 'y' only.

Input: रमा

Output: rama

र-ra म–ma

Input: मनाली

Output: manalee

म-ma न-na ल-la ,ी-ee here ल gives l,' a' is deleted.

**4.2.5 Dealing with Replicative Noun:**

Input: बच्चाबच्चागाँधीजीकोजानताहै

Output: Every child knows Gandhi ji

**4.2.6 Identify singular and plurals**: [4]

If (first word== "मैं"|"तुम"|"वे"|"हम")   Then Word is plural

If (last letter of first word== "या"|"ये"|"यो") Then word is plural

Otherwise it is considered as singular

**4.2.7 Dealing Interrogative Sentences:**

If sentence ends with "ता","ती","ते" Then it is considered as Present indefinite. Word next to "क्या" is checked to identify the number whether it is singular or plural.

Input    क्यासीताखानाखातीहै

Output   Does Sita eats food

In the above example "सीता" is taken as singular so क्या is translated as does.

Input    क्यातुमप्रत्येकमंगलवारफुटबॉलखेलतेहो

Output   Do you play football every Tuesday?

In the above example "तुम" is taken as plural so क्या is translated as do.

if sentence ends with "रहा ","रही","रहे" Then it is considered as continuous sentences. Word next to "क्या" is checked to identify the number whether it is singular or plural.

Input   क्या तुम पढ़ रहे हो

Output: are you reading?

Input    क्या सीता बाज़ार जा रही है

Output   is Sita going to market?

**4.2.8 Dealing with synonyms**:

Input: कृपया ध्यान दे

Output: please pay <u>attention</u>

<u>Meditation</u>

In bilingual dictionary ध्यान is translated as attention and meditation. In our system these synonyms can be handled at runtime.

**4.2.9 Dealing with sentences where exact matching rules are not found:**

Input राम,मोहनऔरश्यामदोस्तहै|रामपुणेमेंरहताहै|

मोहनऔरश्याममुम्बईमेंरहतेहै|

In our system we split the sentences by using "|".Here we get three sentences.

Sentence1:राम,मोहनऔरश्यामदोस्तहै

Sentence2:रामपुणेमेंरहताहै

Sentence3:मोहनऔरश्याममुम्बईमेंरहतेहै

For sentence 1 its corresponding target equivalent structure NOUN NOUNCONJ NOUN AUX NOUN is found in database so it is translated as

Sentence 1: Ram, Mohan and Shyam are friend.

For sentence2, its equivalent target structure is not found in our database. In this case equivalent target structure is searched at phrase level. At phrase level, it is matched with a structure NOUN PREP VERB and its equivalent i.e. VERB PREP NOUN is found in our database .Here it is translated at phrase level.

रामपुणेमेंरहताहै

Ram      lives in Pune

Similarly for sentence3, its equivalent target structure is not found in our database then it is searched at phrase level.

मोहन और श्याम\मुम्बई में रहते है

Mohan and Shyam \ lives in Mumbai

Phrases for which any equivalent target structure is not found, word to word translation is used for them.

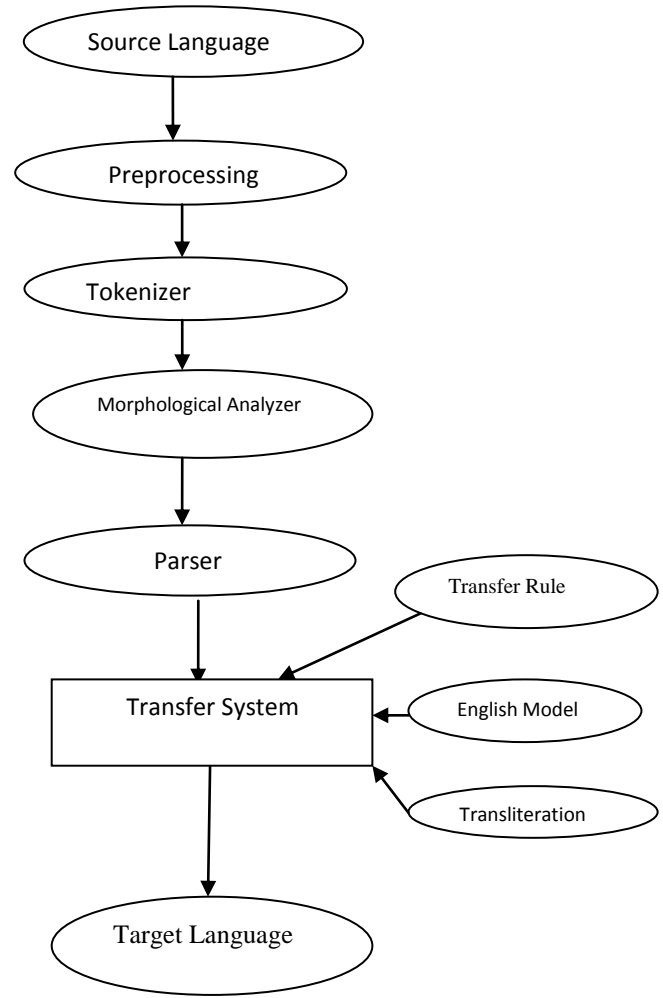

**Fig 4: Flow chart of Transfer Based Machine Translation System**

## 5. Conclusion

In this paper, we have discussed a transfer based machine translation which is based on rule based approach of machine translation. We have incorporated almost all possible translation rules for different Hindi structures. As a future work we can handle the sentences consisting of idioms, complex, compound and other form of sentences in perfect and perfect continuous tenses. We can enhance dictionary to improve translation quality.

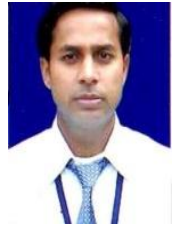

Mr. Shashi Pal Singh is working as STO, AAI Group, C-DAC, Pune. He has completed his B.Tech and M.Tech in Computer Science & Engineering and has published various national & international papers. He is specialized in Natural Language Processing (NLP), Machine assisted Translation (MT), Cloud Computing and   Mobile Computing.

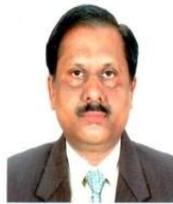

Mr. Ajai Kumar is working as Associate Director and Head, AAI Group, C-DAC, Pune. He is handling various projects in the area of Natural Language Processing, Information Extraction and Retrieval, Intelligent Language Teaching/Tutoring, Speech Technology [Synthesis & Recognition ASR], Mobile Computing, Decision Support Systems & Simulations and has published various national & international papers.

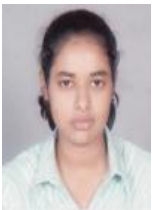

**Ms. Akanksha Gehlot** is pursuing internship from C-DAC Pune under M.Tech, Computer Science 2nd    year Curriculum which is being pursued from Banasthali University, Rajasthan India. She has completed her B.Tech degree in Computer Science from Utter Pradesh Technical University.

akankshagehlot2@gmail.com

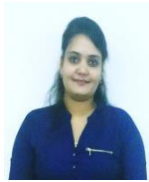

**Ms. Vaishali Sharma** is pursuing internship from C-DAC Pune under M.Tech, Computer Science 2nd    year Curriculum which is being pursued from Banasthali University, Rajasthan India. She has completed her B.Tech degree in Computer Science from Rajasthan Technical University.

vaishali.shrma2217@gmail.com